\documentclass[journal]{IEEEtran}
\usepackage{booktabs}
\usepackage{cite}
\usepackage{graphicx}
\usepackage[cmex10]{amsmath}
\usepackage{latexsym}
\usepackage{amssymb}
\usepackage{mathalfa}
\usepackage{mathrsfs}
\ifCLASSOPTIONcompsoc
\usepackage[caption=false,font=normalsize,labelfont=sf,textfont=sf]{subfig}
\else
  \usepackage[caption=false,font=footnotesize]{subfig}
\fi

\usepackage{fixltx2e}

\begin{document}
\title{Towards the Application of Linear Programming Methods For Multi-Camera Pose Estimation}
\author{Masoud~Aghamohamadian-Sharbaf,~\IEEEmembership{} Ahmadreza~Heravi,~\IEEEmembership{}
    Hamidreza~Pourreza,~\IEEEmembership{Senior Member,~IEEE,}}
%\thanks{Masoud Aghamohamdian-Sharbaf is with the Department of Electrical Engineering, Ferdowsi University of Mashhad, Mashhad, Iran (e-mail: \mbox{masoudas@hotmail.com}).}% <-this % stops a space
%\thanks{Hamidreza Pourreza  is with the Department of Computer Engineering, Ferdowsi University of Mashhad, Mashhad, Iran (e-mail: \mbox{hpourreza@ieee.org})}
%\thanks{Manuscript received ; revised }}
%\markboth{Journal of \LaTeX\ Class Files,~Vol.~11, No.~4, December~2012}%
%{Shell \MakeLowercase{\textit{et al.}}: Bare Demo of IEEEtran.cls for %Journals}
\maketitle

\begin{abstract}

\end{abstract}


\IEEEpeerreviewmaketitle

\section{Introduction}
\IEEEPARstart{C}{amera} pose estimation is a primary requirement for the implementation of vision based multi-camera measurement systems. Pose estimation includes the estimation of six degrees of freedom that expresses the alignment of the camera coordinate system with respect to a universal system. High precision calibration could be performed using three dimensional (3D) \cite{b17,b27} or two dimensional (2D) calibration targets \cite{b2,b18}. However, exposing such objects to all cameras is troublesome. On the other hand, wand-based methods [] which use a 1D object (i.e., a rod with two markers placed at a known distance) overcome such a difficulty and that is why they are frequently used in the state of the art.

To apply wand-based pose estimation method, the wand is waved about in the scene to provide sufficient feature points (FPs) for all cameras. Given that at least two cameras observe the object during the imaging interval, an initial estimation for the marker locations and camera poses is estimated.  Next, these estimation are improved using nonlinear optimization methods. The well-known Levenberg-Marquardt algorithm [] is a powerful tool commonly employed in all previous works and provides reasonable accuracy. However, the algorithm requires a large matrix inversion per iteration which would make it unsuitable if fast calibration is required.

In [], we presented a separation based optimization algorithm which, rather than optimization the entire variables altogether, 
This would allow us to employ: 1) a class of nonlinear functions with three variables and 2) a convex quadratic multi-variable polynomial, for minimization of reprojection error. Neglecting the inversion required to minimize the nonlinear functions, in this paper we demonstrate how separation allows eradication of matrix inversion.  To be more specific, we present a linear programming (LP) solution for the minimization of the convex function. Considering this function contains the majority of variables (which are camera translation and marker locations variables), the proposed optimization method mainly handles pose estimation problem using LP methods (rather than nonlinear optimization methods). Specifically, we employ least absolute error (LAE) rather than least square error (LSE) as a metric for minimization of reprojection error. We demonstrate that if certain (simple) stipulations are imposed on the imaging step, application of LP methods for error minimization of this function would be possible. 

To present our method, in section II we introduce our notation alongside formulating the optimization problem for multi-camera pose estimation. In section III, we briefly view the idea of separation. Next, we discuss how separation allows the optimization problem to be handled using LP methods. In section IV, using simulated and real tests, we demonstrate  the robustness of results and we give the concluding remarks and possible future works in section V.

\section{The Proposed Pose Estimation and Nomenclature}\label{sec:notations-and-outline-of-the-proposed-method} 
\subsection{Proposed Pose Estimation Method}
Throughout this paper, we assume that the internal parameters of all cameras are calibrated using a suitable method such as \cite{}. As stated earlier, wand-based pose estimation constitutes of deriving an initial estimation step and a refinement step based on optimization methods [see Fig.~\ref{}]. A suitable initialization allows the optimization process in the refinement step to converge rapidly. In this paper, we employ a separate static object (depicted in Fig.~\ref{}) to derive an initial estimation for the parameters. Provided that the object is observed by all cameras, which we assume is met in this paper, we use the method of \cite{} to derive an initial estimation for camera poses. The initial poses of the moving object is then estimated using triangulation. In the refinement step, the initial estimations are improved by minimizing the reprojection error of the moving object images. For minimization, we employ the two stage algorithm proposed in \cite{}. The algorithm considers the Euler angles on the one hand and the translation vectors and marker locations on the other hand as independent sets. Then by fixing one set, it estimates the other vice versa and repeats the process until minimal reprojection error is reached. 

Before proceeding with this section, we should discuss a few points regarding the imaging step. The FPs of the calibration objects are extracted using \cite{b34}. Assume that $M'$ images with only two FPs are at hand  for each camera of an $N$ camera constellation. We define a mask matrix $W_{M\times N}$, where $M=2M'$, whose elements are merely one or zero,  If camera $n$ observes the $m$-th 3D point, $w_{mn}$ would be set to one, otherwise it is  zero.
\subsection{Nomenclature}
Throughout this paper, small bold letters denote a vector. Assume that we a assign universal coordinate system to the scene, the marker locations of the moving object in this coordinate system are denoted as $\boldsymbol{x}_m=\begin{bmatrix} x_m & y_m & z_m \end{bmatrix}^T$, $m=1,...,M$.  To model an $N$ camera constellation, we shall express the world points in each camera coordinate system. For this end, the Euler angles and the translation vector of camera $n$ that express the orientation of the camera in the world coordinates are given by $\boldsymbol{\phi}_n=\begin{bmatrix} \phi_{nx} & \phi_{ny} & \phi_{nz} \end{bmatrix}$ and $\boldsymbol{t}_n=\begin{bmatrix} t_{nx}&t_{ny}&t_{nz} \end{bmatrix}^T$, respectively. Moreover, each row of the corresponding rotation matrix $R(\boldsymbol{\phi}_n)=R_z(\phi_{nz}) R_y(\phi_{ny})R_x(\phi_{nx})$ is denoted as $\boldsymbol{r}_{ni}$, $i=1,2,3$, where  $R_x$, $R_y$ and $R_z$ are rotation matrices over the corresponding indexes. Therefore, the relation between a marker location and the corresponding FP is expressed as:
\begin{eqnarray}
\label{pc}
u_{mn}&=&\alpha_{n}+\gamma_n f_n\frac{\boldsymbol{r}_{n1}\boldsymbol{x}_{m}+t'_{nx}}{\boldsymbol{r}_{n3}\boldsymbol{x}_{m}+t'_{nz}},\\ \nonumber v_{mn}&=&\beta_{n}+f_n\frac{\boldsymbol{r}_{n2}\boldsymbol{x}_{m}+t'_{ny}}{\boldsymbol{r}_{n3}\boldsymbol{x}_{m}+t'_{nz}}. 
\end{eqnarray}
where $f_n$,  $\begin{bmatrix}\alpha_{n} & \beta_{n} \end{bmatrix}^T$ and $\gamma_n$ denote the focal length, the principal point and camera aspect ratio for camera $n$, respectively. As a final remark, define the sets of vectors $X$, $T_p$ and $\Phi$ as $X=\{\boldsymbol{x}_m\}_{m=1}^M$, $T_p=\{\boldsymbol{t}_n'\}_{n=1}^N$ and $\Phi=\{\boldsymbol{\phi}_n\}_{n=1}^{N}$.
\begin{figure}[!t] 
\centering
\includegraphics[width=3.5in]{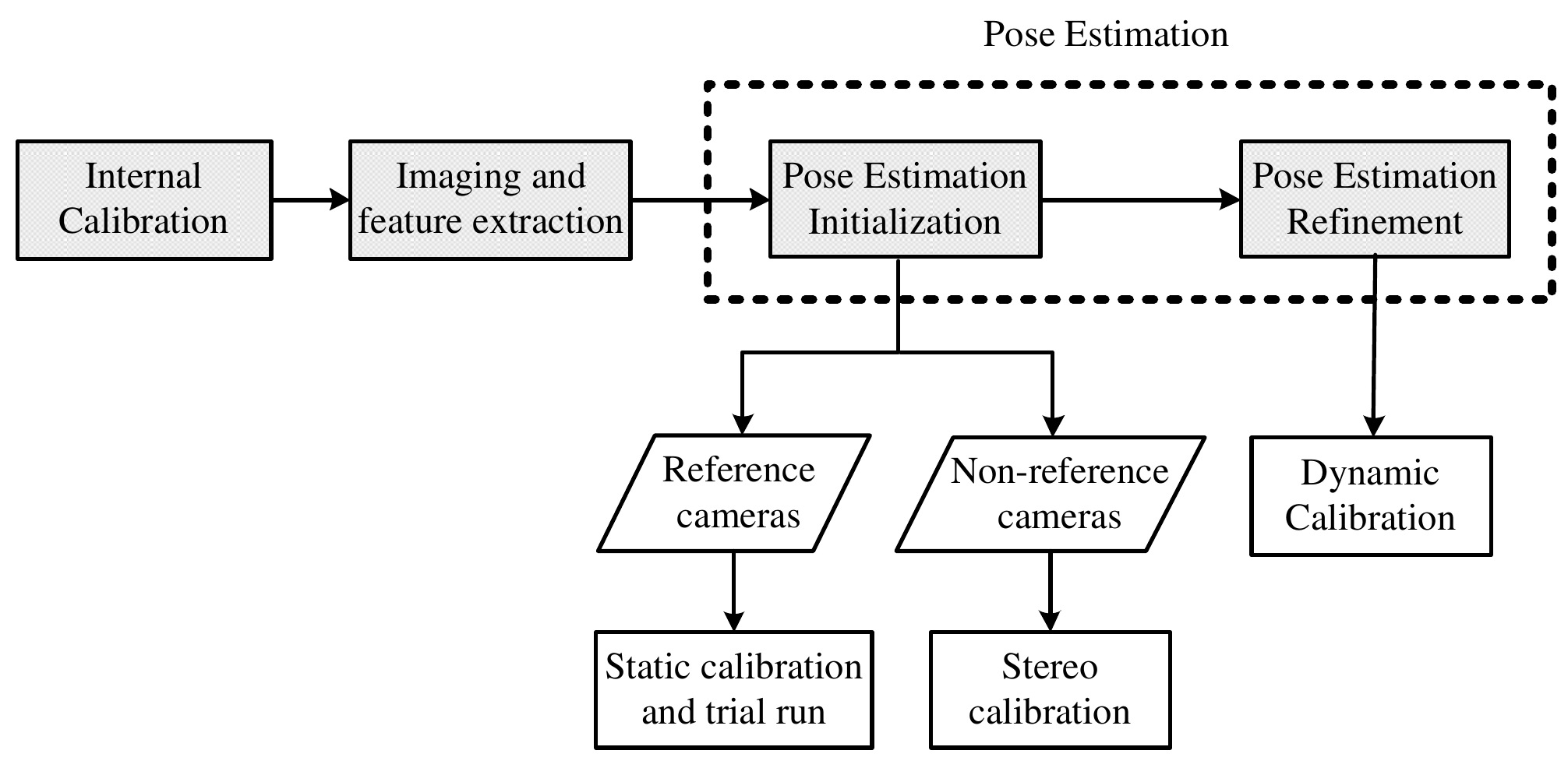}
\caption[Caption for LOF]{Block diagram of the proposed pose estimation algorithm}
\label{fFl}
\end{figure}
\subsection{Refinement Step Formulation}\label{sec:dynamic-calibration}
As stated earlier, the goal of the refinement step in pose estimation is to modify all initial estimations of camera poses and the marker locations so that:
\begin{enumerate}
\item the reprojection error for the set $X$ is minimized,
\item the set $X$ satisfies the constraints imposed by the moving object.
\end{enumerate} 
Therefore, pose estimation refinement could be expressed as an optimization problem. 
To formulate this problem, denote the corresponding FP of $\boldsymbol{x}_m$ in camera $n$ as $\boldsymbol{\tilde{u}}_{mn}=\begin{bmatrix} \tilde{u}_{mn} & \tilde{v}_{mn} \end{bmatrix}^T$, we have:
\begin{align} \label{P}
&\qquad\qquad\qquad\quad\text{min} \quad  P(\Phi,X,T_p) \\ \nonumber
&=\sum_{n=1}^{N}\sum_{m=1}^{M}w_{mn}\Big[(u_{mn}-\tilde{u}_{mn})^2 +(v_{mn}-\tilde{v}_{mn})^2\Big] \\ \nonumber
&\qquad\qquad\text{s.t.}\quad \parallel x_{2m}-x_{2m-1}\parallel^2-d^2=0\quad m=1, ..., M' , 
\end{align}
where $P$ is the reprojection error function in $6N+3M$ variables and $d$ is the distance between the two markers. Note that the a purpose of introducing the length constraint is to ensure that the derived calibrated parameters are in the specified metric for the scene. Moreover, additional constraints  which are introduced when employing more complicated moving objects have no impact on the quality of calibration (see \cite{}). In order to derive the scene scale without constraints, assume that the mean estimated length for the calibration object following the unconstrained optimization is $d_m$. The scene scale is equal to $d/d_m$ and the set $T_p$ should be scaled with this factor to account for the metric of the scene. Henceforth, we eliminate the constraints from the optimization problem.

 It is   possible to further simplify (\ref{P}) by transforming the objective function into a non-fractional from. Define: 
\begin{align}\label{UV}
U_{mn}(\boldsymbol{x}_m,\boldsymbol{t}'_n,\boldsymbol{\phi}_n)=(\bar{u}_{mn}\boldsymbol{r}_{n3}-\boldsymbol{r}_{n1})\boldsymbol{x}_{m}+\bar{u}_{mn}t'_{nz}-t'_{nx} \\ \nonumber
V_{mn}(\boldsymbol{x}_m,\boldsymbol{t}'_n,\boldsymbol{\phi}_n)=(\bar{v}_{mn}\boldsymbol{r}_{n3}-\boldsymbol{r}_{n2})\boldsymbol{x}_{m}+\bar{v}_{mn}t'_{nz}-t'_{ny},
\end{align}
where:
\begin{equation}
\bar{u}_{mn}=\frac{\tilde{u}_{mn}-\alpha_{n}}{f_n}, \quad \bar{v}_{mn}=\frac{\tilde{v}_{mn}-\beta_{n}}{f_n}, \end{equation}
the optimization problem is then expressed by following polynomial like form: 
\begin{align}\label{E}
& \text{min}\quad E(\Phi,X,T_p)=\sum_{n=1}^{N}\sum_{m=1}^{M}w_{mn}\Big[U_{mn}^2+V_{mn}^2\Big]
\end{align}
From this point on, $P$ and $E$ functions denote the functions in (\ref{P}) and (\ref{E}), respectively. It is noteworthy that a translation and/or rotation of all parameters in $E$ and $P$ does not affect the function value. Consequently, both functions possess an infinite number of global minima which further suggests that we arrive at a transformed version of the parameters after optimization. This transformation could be obtained from the deviation of the estimated position for the static object using the refined camera poses.
\section{Optimization of $E$ Based on Separation}
In separation based optimization \cite{} (which is also known as optimization decoupling), the mutual relation between groups of variables in the Taylor series expansion of the function is neglected (i.e., the Hessian matrix is decoupled). Hence, decoupling breaks down  optimization into minimizing  a number of sub-functions derived from the original objective function with fewer variables. Hence, to derive an update for an specific group of variables at each iteration, the sub-function that possesses this group of variables is minimized in the presence of the current estimation for all other variables. This process is then repeated for sufficient number of iterations until the desired minimum is deduced. Such simplification usually enhance optimization in the sense that the computational complexity of each iteration decreases. Although the rate of convergence may decrease due to separation.

Returning to the problem at hand, in \cite{}, we argued that angle variables can be separated from other variables in minimization of $E$. In other words, we argued  that by introducing the following sub-functions:
\begin{equation}\label{Ephi}
E_{\phi_n}=\sum_{m=1}^{M}\left[U^2_{mn}(\boldsymbol{\phi}_n)+V^2_{mn}(\boldsymbol{\phi}_n)\right], \quad n=1, ..., N.
\end{equation}
and :
\begin{equation}\label{Etx}
E_{t,x}=\sum_{n=1}^{N}\sum_{m=1}^{M}\left[U^2_{mn}(\boldsymbol{t}'_n,\boldsymbol{x}_m)+V^2_{mn}(\boldsymbol{t}'_n,\boldsymbol{x}_m)\right]
\end{equation}
the minimization of $E$ could be performed using separation. Under such a scheme for optimization, the family of $N$ functions in (\ref{Ephi}) separately estimate the angles for all cameras. That is, given an estimation $X^{(k-1)}$ and $T_p^{(k-1)}$ derived at iteration $k-1$, $\phi^{(k)}$ could be estimated for all $n$ using LM method as follows:
\begin{equation}
\boldsymbol{\phi}_n^{(k)}=\boldsymbol{\phi}_n^{(k-1)}+\boldsymbol{J}^{T}_{\phi_n}\boldsymbol{J}_{\phi_n}\boldsymbol{J}^{T}_{\phi_n}\Delta \boldsymbol{E}_{\phi_n}
\end{equation}
where $\boldsymbol{J}^{T}_{\phi_n}$ denotes the column Jacobian vector of $E_{\phi_n}$ and $\Delta \boldsymbol{E}_{\phi_n}$ denotes the current error (see \cite{b26}).
The function in () on the other hand is convex quadratic multivariate function with a total of $3M+3N$ variables \cite{}. Given the updated angle vectors, an update for the non-angle variables could be determined by moving along the contours of this function. While Newton method remains a viable approach for such a task, the simple polynomial form of this sub-function motivates us to employ alternative methods.

\subsection{LP minimization of $E_{t,x}$}
The fact that the $U_{mn}$ and $V_{mn}$ are linear polynomials in $X$ and $T$ motivates us to use LP methods for minimization of this problem. As we demonstrate, this is in fact possible if additional linear constrains are imposed on the imaging process. 

In order to eradicate the nonlinearity of the function in (\ref{Etx}), we modify the error metric from LSE to LAE, or:
\begin{equation}\label{Etxabs}
\mathcal{E}_{t,x}=\sum_{m=1}^{M}\sum_{n=1}^{N}w_{mn}\Big[|U_{mn}|+|V_{mn}|\Big]
\end{equation}
Under certain conditions specified in (), the minimum of $E_{t,x}$ and $\mathcal{E}_{t,x}$ are the same. Given that the problem at hand satisfies all such constraints, it would be possible to minimize this function using LP method. For this end, we introduce $2MN$ auxiliary variables $u_{11}, v_{11}, u_{12}, ..., u_{MN}, v_{MN}$  and define the following equivalent problem:
\begin{align} \label{Li}
&\text{min} \hspace{50 pt}  \sum_{m=1}^{M}\sum_{n=1}^{N}w_{mn}\Big[u_{mn}+v_{mn}\Big], \\ \nonumber
&\text{s.t.} \quad  -w_{mn}U_{mn}\leq w_{mn}u_{mn} \quad w_{mn}U_{mn}\leq w_{mn}u_{mn},\quad  \\ \nonumber
&\hspace{22 pt}  -w_{mn}V_{mn}\leq w_{mn}v_{mn} \quad w_{mn}V_{mn}\leq w_{mn}v_{mn},\quad \\ \nonumber
&\hspace{50 pt} B_l \leq t_{nx}, t_{ny}, t_{nz}, x_m, y_m, z_m \leq B_u. 
\end{align}
The above LP is in $2MN+6M+3N$ variables and given that some $w_{mn}$ are zero, it contains at most a total of $4MN$ constraints\footnote{Indeed, we neglect the all zero constraints whilst implementation.} on $U_{mn}$ and $V_{mn}$ and $6M+6N$ extra bound constraints. The role of the constraints on auxiliary variables is to ensure that they remain greater than the absolute values of $U_{mn}$  and $V_{mn}$. Hence, (\ref{Li}) is equivalent to the minimization of (\ref{Etxabs}). As final remark, the bound constraints are set to circumscribe the search space of LP which wound enhance the speed of optimization. Assuming all length units are in meters, we set $B_u=-B_l=10$. This suggests that the cameras and the moving object would not go beyond $10m$ from the origin.

Bearing in mind that it is still impossible to employ LP methods to minimize this subproblem. To elaborate, note that a trivial solution of (\ref{Li}) is:
\begin{equation}\label{Badmin}
\boldsymbol{x}_1=...=\boldsymbol{x}_M=-R^{-1}(\boldsymbol{\phi_1})\boldsymbol{t}'_1=...=-R^{-1}(\boldsymbol{\phi_N})\boldsymbol{t}'_N
\end{equation}
which denotes a solution where all marker locations and translation vectors are assumed equal. These constitute  useless global minima of $\mathcal{E}_{t,x}$ (or $E_{t,x}$) and if not treated properly, cause failure in optimization.  Indeed, when applying the likes of Newton method to minimize $E_{t,x}$ (as we did in \cite{}), the minimization is initiated from the current estimations of $X$ and $T$ variables. Hence, it is unlikely for such method to converge to the unsuitable minima whereas this is an actual possibility for LP methods (which are global optimization method). To remedy this, we need to impose further constraints on $T$ and $X$ variables to fully eradicate such unsuitable minima.

An inspection of (\ref{Badmin}) suggests if additional constraints regarding the relative position of the object and the cameras are added to the problem, this minima could be avoided. For example, if one marker was constantly held above the other, we could write $z_{2m-1}<z_{2m}, m=1, ..., M'$ to eliminate (\ref{Badmin}). However, such restrictions on the movement of the object might be difficult to follow for inexperienced users. A simpler and more practical solution is to  initiate the movement from underneath all cameras (in our case close to the static object). Considering the following two criteria are met:
\begin{enumerate}
\item cameras are placed at higher heights (to fully capture the scene) compared to the static object, and,
\item the frame rate of the cameras is usually high enough to produce many images over a short imaging interval,
\end{enumerate} 
it is guaranteed that the following sets of constraints are satisfied for $m=1, ..., \mathcal{M}$, where $\mathcal{M}\leq M$:
\begin{equation}\label{extra}
w_{mn}z_m\leq -w_{mn}R^{-1}(\boldsymbol{\phi}_n)\boldsymbol{t}_n, \quad n=1, ..., N.
\end{equation}
The above constraints simply imply that the for $\mathcal{M}/2$ frames, the object is at a lower height than all cameras. Given that $M$ is large enough, optimization would not converge to the unsuitable minima and calibration is guaranteed. We found a value of $\mathcal{M}=200$ satisfactory for our implementation, Therefore, we need to add at most $100N$ additional constraints to (\ref{Li}).

In conclusion, to minimize $E$ given a set of initial values for $\Phi$ vectors (such as $\Phi^{(0)}$), we start by minimizing $E_{\phi_n}$ for all $n$. Next, the updated values are fed to \ref{Li} and together with the constraints of \ref{extra}, this problem is minimized. The above process is then repeated for a sufficient number of iterations to deduce the minimum of $E$. An algorithmic implementation of this process is given in Algorithm I (INJa Flow chart ham bad ni).
\subsection{Computational Complexity}

\ifCLASSOPTIONcaptionsoff
  \newpage
\fi

\bibliographystyle{IEEEtran}

\bibliography{IEEEabrv,References}

\end{document}